\documentclass[10pt,twocolumn,letterpaper]{article}

\usepackage[pagenumbers]{cvpr}
\usepackage{amsmath,amssymb,amsfonts}
\usepackage{graphicx}
\usepackage{booktabs}
\usepackage{multirow}
\usepackage{colortbl}
\usepackage{siunitx}
\usepackage{cuted}
\usepackage{caption}
\usepackage{microtype}
\definecolor{cvprblue}{rgb}{0.21,0.49,0.74}
\definecolor{oursrow}{HTML}{EAF4FC}
\definecolor{ourssub}{HTML}{F3F9FD}
\definecolor{refgray}{HTML}{64748B}
\usepackage[pagebackref,breaklinks,colorlinks,allcolors=cvprblue,hyperfootnotes=false]{hyperref}

\renewcommand{\paragraph}[1]{\vspace{0.35em}\noindent\textbf{#1}}

\newcommand{\codeurl}{https://github.com/step-out/NativeSeg}
\newcommand{\weightsurl}{https://huggingface.co/ChengqianMa/NativeSeg}
\newcommand{\blfootnote}[1]{%
  \begingroup
  \renewcommand{\thefootnote}{}%
  \footnotetext{#1}%
  \endgroup
}

\title{NativeSeg: Native Vision Language Segmentation of 3D Computed Tomography via Spatial Visual Tokens and Verified Local Zoom}

\author{
Chengqian Ma$^{1}$\quad
Gaole Dai$^{1}$\quad
Zijie Meng$^{1}$\quad
Weixuan Jin$^{2}$\\[2pt]
Yijie Du$^{3}$\quad
Yuexiao Ma$^{4}$\quad
Xiawu Zheng$^{5}$\quad
Fei Chao$^{5,*}$\\[5pt]
{\normalsize $^{1}$Peking University\quad
$^{2}$Tsinghua University\quad
$^{3}$Institute of Automation, Chinese Academy of Sciences}\\
{\normalsize $^{4}$Nanyang Technological University\quad
$^{5}$Xiamen University}\\[2pt]
{\tt\small machengqian25@stu.pku.edu.cn}
}

\begin{document}
\maketitle

\begin{strip}
    \centering
    \captionsetup{hypcap=false}
    \includegraphics[width=0.95\textwidth]{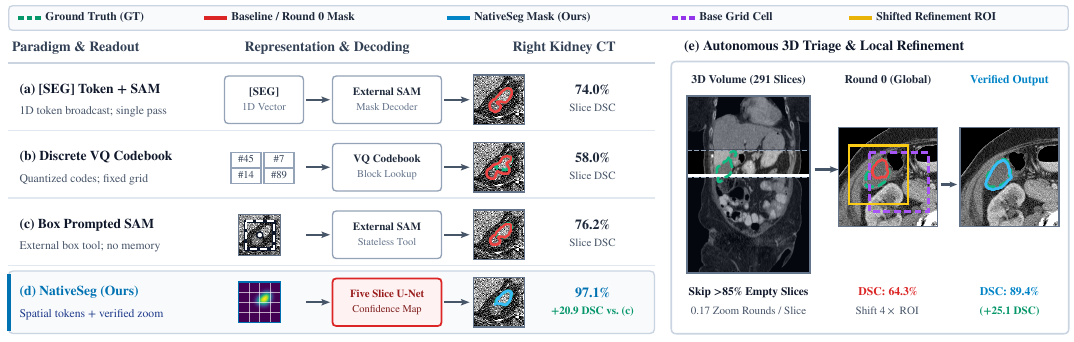}
    \captionof{figure}{Vision language medical segmentation paradigms on an abdominal computed tomography (CT) right kidney slice (a to d) and volumetric triage with verified zoom on adult CT (e). (a) Single pass 1D segmentation token (\texttt{[SEG]}) broadcast into an external Segment Anything Model (SAM) decoder; (b) discrete vector quantized (VQ) codebook lookup; (c) stateless box prompted SAM refinement; (d) \textbf{NativeSeg (Ours)}, decoding confidence maps from 2D spatial tokens in a five slice U-Net decoder with verified region of interest (ROI) zoom to raise slice Dice similarity coefficient (DSC); (e) Round $0$ empty slice screening and gallbladder refinement.}
    \label{fig:paradigms}
    \captionsetup{hypcap=true}
\end{strip}

\begin{abstract}
\blfootnote{$^*$Corresponding author.\\
\textbf{Code:} {\scriptsize\url{\codeurl}}\\
\textbf{Weights:} {\scriptsize\url{\weightsurl}}}%
Text prompted 3D computed tomography (CT) segmentation requires both full slice context and fine boundary precision across volumes where over $80\%$ of organ prompted slices lack the queried target.
Existing medical vision language models (VLMs) are largely \emph{non native} segmenters: they delegate pixel delineation outside the VLM by compressing reasoning into a 1D \texttt{[SEG]} vector for an external Segment Anything Model (SAM) decoder, quantizing masks into lossy discrete codebooks, or issuing box prompts to stateless external SAM tools without empty slice screening or spatial state memory.
We introduce \textbf{NativeSeg}, a \emph{native} vision language segmentation framework on Low Rank Adaptation (LoRA) adapted Qwen3.8-27B that performs dense mask decoding, volumetric triage, spatial error self diagnosis, and multiround local zoom directly over the VLM's 2D spatial visual tokens without external SAM decoders.
NativeSeg decodes confidence maps from upper layer 2D spatial tokens fused with a five slice U-Net pathway, halts at zero zooms on $>91\%$ of standalone evaluations via a native presence gate ($\texttt{<pres>}$), localizes uncertain boundaries via a 20 cell multiscale error gate ($\texttt{<err>}$), and accepts zoomed updates only when predicted error decreases.
Trained solely on adult CT, NativeSeg improves Round $0$ DSC by $+19.81$ and $+7.04$ points over discrete codebook and 1D \texttt{[SEG]} readouts on the identical backbone, reaches $86.81\%$ adult and $87.04\%$ zero shot pediatric DSC, and safely refines 2D nnU-Net ($89.52\%$ adult; $86.86\%$ pediatric), 3D SegVol ($86.81\%$), and 3D VoxTell ($89.85\%$).
\end{abstract}

\section{Introduction}
\label{sec:intro}

Accurate organ delineation in 3D computed tomography (CT) volumes underpins clinical diagnosis, surgical planning, and radiotherapy \cite{isensee2021nnunet,wasserthal2023totalseg}.
Prompting 3D CT segmentation with natural language removes fixed label set constraints, yet exposes a fundamental tension between global anatomical context and fine local boundary resolution: recognizing organ presence and disambiguating adjacent viscera require full slice context across hundreds of axial slices, whereas tracing thin pancreatic tails, gallbladder necks, or pediatric organ walls demands high local spatial resolution \cite{cheng2020cascadepsp,chen2025lrn,pediatricctseg,jordan2022pediatric}.
Moreover, in full volume abdominal evaluation across seven organs, $81\%$ to $86\%$ of organ prompted axial slices do not contain the queried organ \cite{wasserthal2023totalseg}, making uniform high resolution VLM tokenization across every slice quadratically expensive \cite{zheng2026deepeyes,zhang2025acof}.

Although pretrained vision language models (VLMs) offer strong multimodal grounding, current VLM segmentation paradigms remain predominantly \emph{non native}---treating the VLM as a symbolic router that delegates dense spatial perception and refinement to external segmentation machinery (\cref{fig:paradigms}(a to c)).
First, single pass embedding models \cite{lai2024lisa,huang2025medplib,wu2026unibiomed,luo2025vividmed} compress an entire 2D organ shape into a single 1D \texttt{[SEG]} prompt vector fed into an external Segment Anything Model (SAM) mask decoder \cite{kirillov2023sam,ma2024medsam}, discarding the 2D spatial token topology already computed inside the VLM, while discrete codebook models \cite{chang2026medsight,ntpmriseg2026} quantize masks onto coarse block grids.
Second, standard vision towers downsample each CT slice into a $16\times16$ token grid ($>20\,\text{mm}$ per token in adult CT), blurring small structures when decoded in one global pass.
Third, interactive tool calling agents \cite{zhu2025segagent,jiang2026ibisagent,liu2026medsamagent,isensee2026nnintagent} and cascade refiners \cite{chen2025lrn,ma2025medsam2} hand bounding boxes or isolated crops to stateless external segmenters without native empty slice screening, global to local VLM self attention, or closed loop error verification, frequently overwriting accurate contours on clean slices.

We argue that dense perception, volumetric screening, spatial self critique, and local boundary refinement should instead be \emph{native} capabilities of the VLM itself.
We propose \textbf{NativeSeg} (\textbf{Native} Vision Language \textbf{Seg}mentation), a SAM free framework built on Low Rank Adaptation (LoRA) adapted Qwen3.8-27B with a frozen vision tower (\cref{fig:paradigms}(d, e), \cref{fig:overview}).
Rather than routing 1D prompts or box coordinates to external mask heads, NativeSeg operates natively over the VLM's 2D spatial visual tokens across rounds.
In Round $0$, NativeSeg preserves the upper layer $16\times16$ spatial visual token grid and fuses it with a lightweight five slice axial U-Net decoder to predict pixelwise confidence maps directly in original slice coordinates, while a native presence token ($\texttt{<pres>}$) screens out target absent slices.
On candidate slices, without ground truth or external quality estimators, a native multiscale error token ($\texttt{<err>}$) inspects the current mask overlaid on the CT slice to predict misclassified pixel area across $1\times$ global, $2\times2$ quadrant, and $4\times4$ fine regions, halting immediately when predicted error is small or zooming into uncertain boundaries.
During each zoom round, high resolution local visual tokens attend natively to full slice global visual tokens inside the VLM, and candidate edits are accepted only if $\texttt{<err>}$ verifies that predicted error decreases both locally and globally---supporting both \textbf{standalone volumetric segmentation} from scratch and \textbf{plug and play refinement} of external 2D or 3D specialists.

Evaluated across complete 3D CT volumes without slice prefiltering on adult TotalSegmentator \cite{wasserthal2023totalseg} and external zero shot Pediatric-CT-SEG \cite{pediatricctseg,jordan2022pediatric}, this work makes three contributions:
\begin{itemize}
\item \textbf{Native 2D Spatial Token Readout without External SAM Heads.} We show that preserving the VLM's native 2D spatial visual token grid and fusing it with local five slice CT context eliminates the geometric bottleneck of 1D \texttt{[SEG]} prompts and discrete mask codebooks. On the identical Qwen3.8-27B backbone and adult training split, native Round $0$ readout improves volumetric Dice similarity coefficient (DSC) by $+19.81$ and $+7.04$ points over discrete codebook and 1D \texttt{[SEG]} broadcast baselines (\cref{tab:ablations}(a)).
\item \textbf{Native Volumetric Screening and Ground Truth Free Error Localization.} By embedding learnable $\texttt{<pres>}$ and 20 cell multiscale $\texttt{<err>}$ tokens directly into the VLM forward pass over visual mask overlays, NativeSeg screens empty slices and ranks local boundary errors natively without external critic networks, halting at zero zooms on $>91\%$ of full volume standalone evaluations ($0.14$ to $0.17$ mean executed zooms/slice; \cref{fig:error_pareto}).
\item \textbf{Verified Native Local Zoom Across Standalone and Specialist Refiner Modes.} Coupling joint global and local VLM attention with closed loop error verification and 3D axial continuity raises standalone accuracy to $86.81\%$ adult test DSC ($p=2.3\times10^{-7}$) and $87.04\%$ zero shot pediatric DSC ($p=6.7\times10^{-6}$), while safely improving 2D nnU-Net ($89.52\%$ adult; $86.86\%$ pediatric, $p=0.024$), 3D SegVol ($86.81\%$, $p=9.5\times10^{-6}$), and 3D VoxTell ($89.85\%$) where external crop refiners degrade base masks (\Cref{tab:ts_main,tab:pediatric}).
\end{itemize}

\section{Related Work}
\label{sec:related}

\paragraph{Text Prompted Medical Image Segmentation.}
Natural language prompting enables flexible anatomical segmentation across imaging modalities \cite{liu2023clip,zhao2025biomedparse,du2024segvol,huang2024cat,zhao2025sat,li2026openvocabct,rokuss2026voxtell}.
Volumetric specialist and foundation models, including the CLIP-Driven Universal Model \cite{liu2023clip}, CAT \cite{huang2024cat}, SegVol \cite{du2024segvol}, SAT \cite{zhao2025sat}, OpenVocabCT \cite{li2026openvocabct}, and VoxTell \cite{rokuss2026voxtell}, pair text encoders with dedicated 3D U-Net or 3D Vision Transformer (ViT) backbones, while Medal~S \cite{shi2025medals} and ESICA \cite{xin2026esica} incorporate first stage 3D predictions as spatial prompts at fixed voxel resolution.
Complementarily, 2D foundation models such as BiomedParse \cite{zhao2025biomedparse} and BoltzFormer \cite{zhao2025boltzmann} process individual slices at a single global scale.
NativeSeg bridges general purpose VLMs with volumetric CT segmentation by keeping dense decoding, Round $0$ empty slice screening, and selective local zooming native to the VLM while also serving as a plug and play refiner for 2D and 3D segmenters.

\paragraph{From Extrinsic Mask Heads to Native VLM Segmentation.}
Existing VLM segmentation architectures follow three designs that differ in whether spatial perception is delegated outside or kept native inside the VLM.
First, building on LISA \cite{lai2024lisa}, biomedical VLMs including MedPLIB \cite{huang2025medplib}, VividMed \cite{luo2025vividmed}, Citrus-V \cite{wang2025citrusv}, UniBiomed \cite{wu2026unibiomed}, and ROSALIA \cite{choi2026rosalia} collapse multimodal reasoning into a 1D \texttt{[SEG]} token embedding and delegate pixel delineation to an external SAM mask decoder \cite{kirillov2023sam,ma2024medsam}.
Second, Text4Seg \cite{lan2025text4seg}, MedSIGHT \cite{chang2026medsight}, and NTP-MRISeg \cite{ntpmriseg2026} quantize masks into discrete codebook tokens within the large language model (LLM) vocabulary for autoregressive generation.
Third, natural image methods such as F-LMM \cite{wu2025flmm}, Pixel-SAIL \cite{zhang2025pixelsail}, MLLMSeg \cite{wang2025mllmseg}, and DenseMLLM \cite{li2026densemllm} read masks natively from 2D spatial visual tokens in a single global pass, while Liu \etal \cite{liu2026ctreasoning} apply user supplied ROI crops for CT question answering.
NativeSeg eliminates both external SAM decoders and discrete block quantization, extending native 2D spatial token readout \cite{wu2025flmm,li2026densemllm} beyond its coarse $16\times16$ global token grid through five slice axial conditioning and autonomous $2\times$ to $4\times$ local reencoding.

\paragraph{Tool Calling Agents vs.\ Native Closed Loop Refinement.}
Multiturn segmentation agents including SegAgent \cite{zhu2025segagent}, IBISAgent \cite{jiang2026ibisagent}, Ophiuchus \cite{jiang2026ophiuchus}, MedSAM-Agent \cite{liu2026medsamagent}, XFlow \cite{choi2026xflow}, and nnInteractive Agent \cite{isensee2025nninteractive,isensee2026nnintagent} iteratively emit point, scribble, or box prompts to a stateless external segmenter, or adjust parametric contours as in LLMSeg \cite{hamdi2026llmseg}.
In parallel, cascade postprocessors \cite{cheng2020cascadepsp,wang2023segrefiner,chen2025lrn,delanglais2025score}, segmentation quality estimators \cite{zenk2025failure,specktor2025segqc,qiu2026qcresunet,chen2026segae,sheng2025hcr}, and adaptive visual zoom policies \cite{banino2021pondernet,zheng2026deepeyes,zhang2025acof} train separate networks for error estimation or crop editing.
Rather than handing symbolic prompts to an external crop tool without full slice VLM context or acceptance checks, NativeSeg performs error localization ($\texttt{<err>}$), global to local visual attention, and closed loop acceptance verification natively inside the VLM.

\section{Method}
\label{sec:method}

\begin{figure*}[t]
\centering
\includegraphics[width=0.94\textwidth]{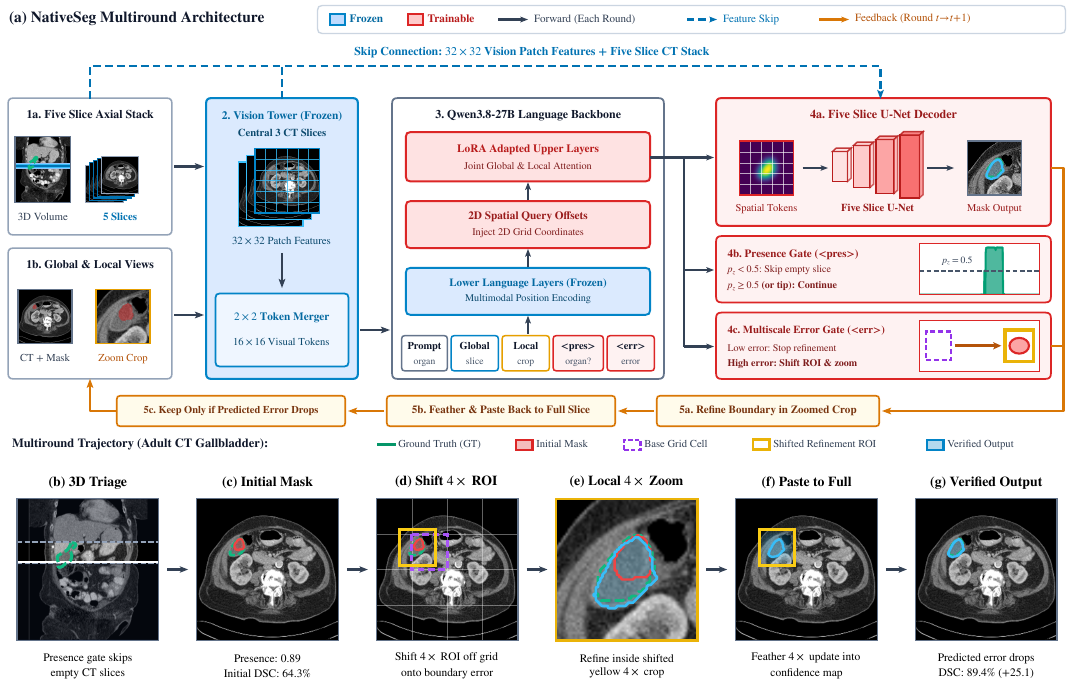}
\caption{Architecture and native volumetric pipeline of \textbf{NativeSeg}. (a) Given the five slice stack (1a) and global or zoomed overlay views (1b), the frozen vision tower (2) encodes CT views while the LoRA adapted Qwen3.8-27B backbone (3) processes five native token groups without external SAM heads: organ \textbf{Prompt} ($q$), \textbf{Global} and \textbf{Local} 2D spatial visual tokens, $\texttt{<pres>}$, and $\texttt{<err>}$. Upper layer states drive dense spatial readout (4a), empty slice presence screening (4b), and 20 cell multiscale error diagnosis (4c) to halt when predicted error is small (\cref{eq:stop}) or zoom into uncertain boundaries (\cref{eq:shift}) with closed loop verification (Steps 5a to 5c). (b to g) Multiround gallbladder refinement example.}
\label{fig:overview}
\end{figure*}

\subsection{Native Multiround Formulation and Multiscale Views}
Given an axial CT slice $I_z\in\mathbb{R}^{H\times W}$ at axial index $z$ (with original slice height $H$ and width $W$, paired with adjacent slices) and a text prompt $q$ (e.g., ``Segment the pancreas.''), non native VLM segmenters compress multimodal context into a 1D \texttt{[SEG]} vector or box coordinates for an external SAM head.
Instead, we formulate segmentation as an autonomous multiround spatial state evolution kept native to the VLM:
\begin{equation}
(L_0, M_0) \to (L_1, M_1) \to \dots \to (L_T, M_T), \quad \hat{M} = M_T,
\label{eq:loop}
\end{equation}
where $L_t\in\mathbb{R}^{H\times W}$ denotes pixelwise log odds (logits) at round $t$, $S_t = \sigma(L_t)\in[0,1]^{H\times W}$ is the corresponding probability map under the logistic sigmoid $\sigma(\cdot)$, $M_t = \mathbb{1}[L_t > 0]\in\{0,1\}^{H\times W}$ is the binary mask ($\mathbb{1}[\cdot]$ denotes the indicator function), $\hat{M}$ is the slice output prior to 3D assembly, $K$ is the maximum zoom budget, and $T \in \{0, \dots, K\}$ is the adaptive stopping round (\cref{fig:overview}).
\Cref{eq:loop} supports two inference modes without external mask tools: \textbf{standalone mode} decodes Round $0$ proposals from scratch, whereas \textbf{plug and play refiner mode} initializes $M_0 = M_0^{\text{spec}}$ from any external 2D or 3D segmenter and refines uncertain slices via Steps 2 to 4.
Because over $80\%$ of organ prompted CT slices lack the queried organ, a native presence gate ($\texttt{<pres>}$) screens empty slices at $T=0$ to avoid wasted zooms and empty slice false positives.

To standardize soft tissue contrast, CT intensities in Hounsfield units (HU) are clipped to $[-160,240]\,\text{HU}$, normalized to $[0,1]$, and letterboxed onto a $512\times512$ canvas ($s_{\text{canvas}} = 512/\max(H,W)$) to form the global view $V_0$.
As shown in \cref{fig:overview}(a), the frozen Qwen3.8-27B vision tower partitions any $512\times512$ view into a $32\times32$ patch grid to extract patch features $F_{V_0}$, and merges $2\times2$ neighboring patches into $256$ 2D spatial visual tokens $X_0\in\mathbb{R}^{256\times D}$ on a $16\times16$ grid ($D$ denotes the language hidden dimension; $32\times32$ canvas pixels per token).
To match errors across organ scales, we define $20$ multiscale candidate grid cells alongside the full canvas ($c=0$): $4$ quadrants ($2\times2$ grid, $c\in\{1,\dots,4\}$, $2\times$ linear zoom) and $16$ fine cells ($4\times4$ grid, $c\in\{5,\dots,20\}$, $4\times$ linear zoom), optionally shifting the local ROI $r^\star$ continuously toward adjacent high error cells (\cref{eq:shift}).

\subsection{Native VLM Readout, Self Diagnosis, and Local Zoom}
All modules build natively on the Qwen3.8-27B language backbone equipped with rank $16$ LoRA adapters \cite{hu2022lora} ($\alpha_{\text{LoRA}}=32$) on the top $16$ layers (Layers $49$ to $64$), keeping the vision tower and lower $48$ layers frozen.
Across rounds, the backbone processes up to five native token groups (\cref{fig:overview}(a), Stage 3): the text \textbf{Prompt} $q$, full slice \textbf{Global} visual tokens, zoomed \textbf{Local} visual tokens, a learnable presence token $\texttt{<pres>}$, and a learnable error token $\texttt{<err>}$.
Placeholder tokens traverse the frozen lower $48$ layers, and learnable task embeddings plus a 2D spatial query offset table $o_{\text{sp}}\in\mathbb{R}^{256\times D}$ (with distinct global/zoom type embeddings) enter the residual stream at Layer $49$, avoiding backward passes through Layers $1$ to $48$.

\paragraph{Step 1 (Round 0): Native 2D Spatial Token Readout and Presence Gate.}
In Round $0$, given the five slice axial stack $[I_{z-2},\dots,I_{z+2}]$, the frozen vision tower encodes the central three slices $[I_{z-1}, I_z, I_{z+1}]$ stacked across RGB channels as $V_0\in[0,1]^{3\times512\times512}$, yielding $32\times32$ patch features $F_{V_0}$ and $256$ visual tokens $X_0$.
Processing $[q;\,X_0;\,\texttt{<pres>}]$ inside the VLM, we extract the Layer $64$ 2D spatial token states $H_{V_0}\in\mathbb{R}^{256\times D}$, reshape them to $16\times16$, upsample to $32\times32$, and fuse them with $F_{V_0}$ (yielding a lightweight single slice spatial decoder head, or our full five slice U-Net decoder).
To preserve subtle parenchymal boundaries across slices, our five slice U-Net decoder (a five stage residual encoder decoder on the $512\times512$ canvas) encodes $[I_{z-2},\dots,I_{z+2}]$ alongside a narrow soft tissue window $[-60,140]\,\text{HU}$, modulates its $32\times32$ bottleneck with the VLM spatial features, and maps the output back to original $(H,W)$ slice coordinates as raw Round $0$ logits $\tilde{L}_{0,z}$ ($B_z = \mathbb{1}[\tilde{L}_{0,z} > 0]$), while a linear head on $\texttt{<pres>}$ predicts organ presence probability $p_z \in [0, 1]$ (halting immediately at $T=0$ when the Round $0$ mask is empty and $p_z < 0.1$).

\paragraph{Step 2: Native Multiscale Error Gate and Freeform ROI Shift.}
At test time (without ground truth), the current mask $M_t$ is rendered onto the $512\times512$ central CT slice with translucent red fill and contour opacity $\alpha_{\text{ov}}=0.35$ to form the global overlay view $O_0^{(t)}$ (visual tokens $X_{O_0}^{(t)}$).
Processing $[q;\,X_{O_0}^{(t)};\,\texttt{<err>}]$ natively inside the VLM, a linear head on $\texttt{<err>}$ outputs raw log error scores $\hat{z}_c^{(t)}$ for the full canvas ($c=0$) and all $20$ candidate cells ($c\in\{1,\dots,20\}$), mapped to non negative predicted misclassified canvas pixel areas $E_c^{(t)} = \exp(\max(\hat{z}_c^{(t)}, 0)) - 1 \ge 0$.
Because $2\times2$ quadrants span $4\times$ the area of $4\times4$ cells, broad multi cell errors naturally select a $2\times$ quadrant first, whereas compact errors or later rounds (with visited cells tracked in $\mathcal{V}_t$) target $4\times4$ fine cells.
Refinement halts ($T=t$) when the highest predicted error among unvisited cells falls below tolerance:
\begin{equation}
E_{c^\star}^{(t)} < \tau \max\bigl(s_{\text{canvas}}^2 |M_t|,\, a_{\min}\bigr),
\label{eq:stop}
\end{equation}
where $c^\star = \arg\max_{c\in\{1,\dots,20\}\setminus\mathcal{V}_t} E_c^{(t)}$, $\tau > 0$ is the relative error tolerance, $|M_t| = \sum_x M_t(x)$, and $a_{\min}=800$ canvas pixels avoids spurious zooms on tiny masks.
When freeform ROI shifting is enabled, the crop center $\hat{p} = (\hat{v}, \hat{u})\in[0,512]^2$ (side length $s_{c^\star}\in\{256,128\}$) shifts from cell center $p_{c^\star}$ toward 8-connected same scale neighbors $\mathcal{N}_+ = \{c^\star\}\cup\mathcal{N}(c^\star)$ exceeding $0.40\,E_{c^\star}^{(t)}$ ($0.45$ for quadrants, $\eta=0.35$):
\begin{equation}
\hat{p} = (1 - \eta)\,p_{c^\star} + \eta\,\frac{\sum_{j\in\mathcal{N}_+} \bigl(E_j^{(t)}\bigr)^2 p_j}{\sum_{j\in\mathcal{N}_+} \bigl(E_j^{(t)}\bigr)^2}.
\label{eq:shift}
\end{equation}

\begin{figure*}[t]
\centering
\includegraphics[width=0.80\textwidth]{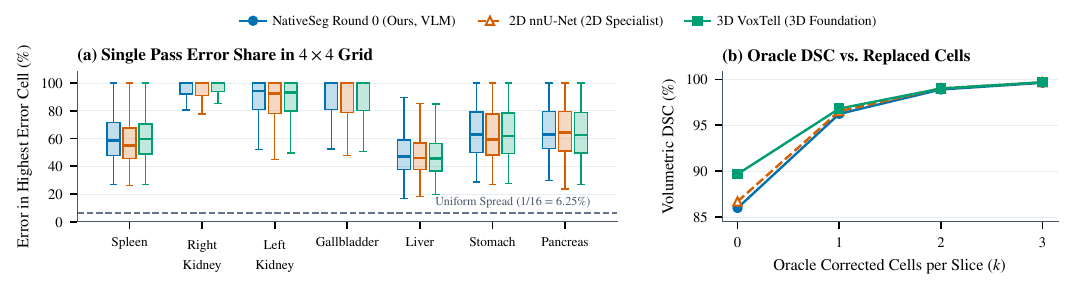}
\caption{Post hoc diagnostic analysis of single pass error concentration on target present Pediatric-CT-SEG slices across \textbf{NativeSeg Round $0$} (blue), \textbf{2D nnU-Net} (vermilion), and \textbf{3D VoxTell} (green). (a) Share of misclassified pixels in the highest error cell of a $4\times4$ grid ($6.25\%$ uniform baseline, dashed; boxes: interquartile range and median; whiskers: $1.5\times$ interquartile range). (b) Oracle 3D DSC ($\uparrow$) after replacing $k\in\{1,2,3\}$ highest error $4\times4$ cells per slice with ground truth ($k=0$: single pass).}
\label{fig:locality}
\end{figure*}

Post hoc analysis on frozen single pass predictions (\cref{fig:locality}, unused for tuning) motivates local zooming: whereas uniform scatter places $6.25\%$ of wrong pixels per $4\times4$ cell, the single worst cell captures a median of $65.9\%$ to $67.7\%$ of total slice error across all three models (\cref{fig:locality}(a), over $10\times$ uniform density), and oracle replacement of $k=1$ cell reaches $96.24\%$ to $96.81\%$ volumetric DSC (\cref{fig:locality}(b)).

\paragraph{Step 3 (Round $t+1$): Native Global and Local Self Attention and State Update.}
For selected region $r^\star$, we crop and resize the local CT view $V_{r^\star}$, local overlay view $O_{r^\star}^{(t)}$, and local prior state $S_{r^\star}^{(t)}\in[0,1]^{512\times512}$ to $512\times512$ ($2\times$ or $4\times$ higher linear sampling resolution), and process the joint sequence natively inside the VLM:
\begin{equation}
\bigl[q;\, X_{O_0}^{(t)}\text{ (global context)};\, X_{O_{r^\star}}^{(t)}\text{ (local detail)}\bigr],
\label{eq:seq}
\end{equation}
allowing every zoomed local visual token in $X_{O_{r^\star}}^{(t)}$ to attend to $q$ and the full slice state $X_{O_0}^{(t)}$ across Layers $49$ to $64$.
The spatial decoder is anchored to the prior state $S_{r^\star}^{(t)}$ via a residual skip:
\begin{align}
L_{r^\star} ={}& \mathrm{Decoder}\bigl(H_{O_{r^\star}}, F_{V_{r^\star}}, V_{r^\star}, S_{r^\star}^{(t)}\bigr) \nonumber \\
& + \lambda_{\text{res}}\bigl(2\,S_{r^\star}^{(t)} - 1\bigr),
\label{eq:residual}
\end{align}
where $H_{O_{r^\star}}$ denotes Layer $64$ spatial states of $O_{r^\star}^{(t)}$, $F_{V_{r^\star}}$ denotes vision patch features of $V_{r^\star}$, and $\lambda_{\text{res}} > 0$ weights the $[-1,1]$ prior anchor.
Mapped back to original $(H,W)$ coordinates as $\hat{L}_{r^\star}$, updates merge into $L_t$ via window $W_{r^\star}(x)\in[0,1]$ ($1$ inside $r^\star$, with optional $6\%$ raised cosine border feathering when shifting freeform ROIs; \cref{sec:supp_repro}):
\begin{equation}
\tilde{L}_{t+1}(x) = \bigl(1 - W_{r^\star}(x)\bigr) L_t(x) + W_{r^\star}(x)\,\hat{L}_{r^\star}(x),
\label{eq:blend}
\end{equation}
yielding candidate $(\tilde{L}_{t+1}, \tilde{M}_{t+1})$ with $\tilde{M}_{t+1} = \mathbb{1}[\tilde{L}_{t+1} > 0]$.

\paragraph{Step 4: Closed Loop Verification, Volumetric Continuity, and Consensus.}
Rendering $\tilde{M}_{t+1}$ into candidate global overlay $\tilde{O}_0^{(t+1)}$, we query $\texttt{<err>}$ for candidate errors $\tilde{E}_c^{(t+1)}$ and accept the update if and only if predicted error drops locally without growing globally:
\begin{equation}
\left\{
\begin{aligned}
\tilde{E}_{c^\star}^{(t+1)} &< E_{c^\star}^{(t)}, \\
\tilde{E}_0^{(t+1)} &\le \min\bigl(E_0^{(t)},\, E_0^{(0)}\bigr),
\end{aligned}
\right.
\label{eq:verify}
\end{equation}
reverting to $(L_t, M_t, E^{(t)})$ otherwise while marking $\mathcal{V}_{t+1} = \mathcal{V}_t \cup \{c^\star\}$.
When assembling 3D standalone volumes across slices, deterministic axial continuity and 3D component filtering (\cref{sec:supp_repro}) retain high confidence core slices, recover overlapping adjacent organ tips ($p_z \ge 0.25$) and single slice axial gaps, prune unsupported isolated slices ($p_z < 0.62$), and filter spurious 3D islands whose size weighted presence falls below $0.22$ of the dominant organ component.
In specialist refiner mode on $M_0^{\text{spec}}$, cross model consensus retains verified boundary edits supported by standalone Round $0$ ($p_z \ge 0.5$ for foreground additions) and prunes isolated single slice specialist false positives when $z\pm 1$ are empty and $p_z < 0.03$.

\subsection{Curriculum Training and Modular Execution}
In \textbf{Stage A} (Round $0$ readout), we optimize the top 16 layer LoRA adapters, spatial offsets, presence head, and spatial decoder on organ balanced slices:
\begin{align}
\mathcal{L}_{\text{init}} ={}& \mathcal{L}_{\text{BCE}}^{\text{logit}}(\tilde{L}_{0,z}, Y)
+ a_z\,\mathcal{L}_{\text{Dice}}\bigl(\sigma(\tilde{L}_{0,z}), Y\bigr) \nonumber \\
& + 0.5\,\mathcal{L}_{\text{BCE}}^{\text{prob}}(p_z, a_z),
\label{eq:loss}
\end{align}
where $Y\in\{0,1\}^{H\times W}$ is the ground truth mask and $a_z = \mathbb{1}[|Y| > 0]$.
In \textbf{Stage B} (multiround error gating and local zoom), we jointly optimize $\mathcal{L}_{\text{init}}$, 21 region smooth $L_1$ log error regression against $\log(1+e_c^{(t)})$ ($e_c^{(t)} = s_{\text{canvas}}^2\sum_{x\in\Omega_c}\mathbb{1}[M_t(x)\neq Y(x)]$), and boundary weighted zoom loss ($3\times$ to $3.5\times$ logit BCE weight on misclassified pixels $M_t(x)\neq Y(x)$ plus soft Dice across clean, perturbed, Round $0$, and multi zoom mask states; \cref{sec:supp_repro}).
At inference, NativeSeg supports both a single checkpoint joint five slice Stage B model (\cref{tab:ablations}(b)) and a modular stage decoupled pipeline that pairs five slice Round $0$ readout with our dedicated single slice local zoom refiner across full 3D benchmarks (\Cref{tab:ts_main,tab:pediatric,tab:ts_val}; \cref{sec:supp_repro}).

\section{Experimental Setup}
\label{sec:setup}

\begin{table*}[t]
\centering
\scriptsize
\renewcommand{\arraystretch}{0.74}
\setlength{\tabcolsep}{3.5pt}
\caption{In domain segmentation and specialist refinement on the TotalSegmentator test split (65 patients, 20,506 axial slices, 143,542 organ slice evaluations; $^\dagger$controlled readout paradigm adapted to Qwen3.8-27B or TotalSegmentator perturbation trained LRN reimplementation; organ columns report 3D Dice similarity coefficient [DSC, \%]; NSD: normalized surface Dice at 3\,mm [\%]; HD95: 95th percentile Hausdorff distance [mm]; Miss / FP: missed positive / false positive empty slice evaluations; Spl: spleen, RKid/LKid: right/left kidney, Gall: gallbladder, Liv: liver, Sto: stomach, Pan: pancreas).}
\label{tab:ts_main}
\begin{tabular}{lrrrrrrrrrrr}
\toprule
Method & Spl $\uparrow$ & RKid $\uparrow$ & LKid $\uparrow$ & Gall $\uparrow$ & Liv $\uparrow$ & Sto $\uparrow$ & Pan $\uparrow$ & Mean DSC (\%) $\uparrow$ & NSD (\%) $\uparrow$ & HD95 (mm) $\downarrow$ & Miss $\downarrow$ / FP $\downarrow$ \\
\midrule
\multicolumn{12}{l}{\textit{(a) Standalone VLM segmentation (all models trained on the 1,081 adult CT scans of TotalSegmentator)}} \\
\midrule
LISA$^\dagger$ \cite{lai2024lisa} & 93.40 & 86.03 & 86.04 & 70.84 & 91.67 & 73.96 & 59.79 & 80.25 & 76.37 & 21.24 & 1483 / 1134 \\
DenseMLLM$^\dagger$ \cite{li2026densemllm} & 94.22 & 87.37 & 86.87 & 72.60 & 92.38 & 78.90 & 63.18 & 82.22 & 79.47 & 19.26 & 1304 / 957 \\
F-LMM$^\dagger$ \cite{wu2025flmm} & 95.85 & 88.30 & 87.11 & 75.42 & 93.34 & 83.62 & 70.19 & 84.83 & 83.79 & 14.72 & 1195 / 879 \\
Single Slice Multiround VLM (Ours) & 96.13 & 88.62 & 87.47 & \textbf{78.40} & 93.43 & 85.34 & 71.11 & 85.78 & 86.25 & 8.93 & 1022 / \textbf{533} \\
NativeSeg Round 0 (Ours) & 96.52 & \textbf{89.63} & 87.19 & 75.21 & 93.51 & 85.77 & 74.22 & 86.01 & 85.76 & 13.13 & 915 / 801 \\
\rowcolor{oursrow}\textbf{NativeSeg (Ours)} & \textbf{96.71} & 89.30 & \textbf{87.95} & 78.39 & \textbf{93.61} & \textbf{86.79} & \textbf{74.91} & \textbf{86.81} & \textbf{87.66} & \textbf{7.87} & \textbf{711} / 550 \\
\midrule
\multicolumn{12}{l}{\textit{(b) Plug and play refinement of 2D nnU-Net (where 2D nnU-Net is trained on the same 1,081 adult CT scans)}} \\
\midrule
2D nnU-Net \cite{isensee2021nnunet} & 97.39 & 91.20 & 92.53 & 80.08 & \textbf{94.78} & \textbf{90.24} & \textbf{79.99} & 89.46 & 91.73 & 5.80 & 529 / 482 \\
2D nnU-Net + LRN$^\dagger$ \cite{chen2025lrn} & 94.13 & 83.61 & 79.44 & 58.04 & 94.33 & 82.74 & 64.53 & 79.55 & 66.30 & 207.63 & \textbf{45} / 44799 \\
2D nnU-Net + MedSAM2 \cite{ma2025medsam2} & 96.49 & 89.52 & 91.49 & 78.74 & 92.63 & 87.15 & 75.56 & 87.37 & 88.66 & 8.32 & 529 / 480 \\
\rowcolor{ourssub}\textbf{2D nnU-Net + NativeSeg (Ours)} & \textbf{97.42} & \textbf{91.30} & \textbf{92.75} & \textbf{80.26} & 94.77 & 90.17 & 79.96 & \textbf{89.52} & \textbf{91.76} & \textbf{5.74} & 529 / \textbf{471} \\
\bottomrule
\end{tabular}
\end{table*}

\begin{table*}[t]
\centering
\scriptsize
\renewcommand{\arraystretch}{0.68}
\setlength{\tabcolsep}{3.5pt}
\newcommand{\rg}[1]{\textcolor{refgray}{#1}}
\caption{Zero shot generalization on external Pediatric-CT-SEG (21 patients, 5,366 axial slices, 37,562 organ slice evaluations [7,102 positive, 30,460 absent]; gray reference block lists external multi dataset foundation models trained on large scale extra corpora [$^*$reported in \cite{rokuss2026voxtell}; $^\ddagger$official checkpoint evaluated across all 37,562 evaluations]; $^\dagger$controlled Qwen3.8-27B VLM or LRN reimplementation trained solely on 1,081 TotalSegmentator adult CT scans; Spl: spleen, RKid/LKid: right/left kidney, Gall: gallbladder, Liv: liver, Sto: stomach, Pan: pancreas).}
\label{tab:pediatric}
\begin{tabular}{lrrrrrrrrrrr}
\toprule
Method & Spl $\uparrow$ & RKid $\uparrow$ & LKid $\uparrow$ & Gall $\uparrow$ & Liv $\uparrow$ & Sto $\uparrow$ & Pan $\uparrow$ & Mean DSC (\%) $\uparrow$ & NSD (\%) $\uparrow$ & HD95 (mm) $\downarrow$ & Miss $\downarrow$ / FP $\downarrow$ \\
\midrule
\multicolumn{12}{l}{\rg{\textit{Reference: External multi dataset 2D/3D foundation models trained on large scale extra corpora (up to 22K+ volumes; not TotalSeg only)}}} \\
\midrule
\rg{BiomedParse$^*$} \cite{zhao2025biomedparse} & \rg{36.20} & \rg{18.77} & \rg{22.86} & \rg{8.74} & \rg{35.62} & \rg{14.47} & \rg{11.66} & \rg{21.19} & \rg{Not rep.} & \rg{Not rep.} & \rg{Not rep.} \\
\rg{Text3DSAM$^*$} \cite{xin2026text3dsam} & \rg{64.27} & \rg{28.35} & \rg{59.28} & \rg{7.25} & \rg{76.64} & \rg{48.37} & \rg{14.30} & \rg{42.64} & \rg{Not rep.} & \rg{Not rep.} & \rg{Not rep.} \\
\rg{3D SegVol$^\ddagger$} \cite{du2024segvol} & \rg{88.95} & \rg{89.00} & \rg{92.46} & \rg{65.22} & \rg{95.75} & \rg{86.13} & \rg{76.89} & \rg{84.91} & \rg{86.01} & \rg{10.05} & \rg{720 / 167} \\
\rg{3D SAT$^*$} \cite{zhao2025sat} & \rg{93.10} & \rg{94.26} & \rg{96.12} & \rg{80.25} & \rg{95.65} & \rg{77.87} & \rg{78.81} & \rg{88.01} & \rg{Not rep.} & \rg{Not rep.} & \rg{Not rep.} \\
\rg{BiomedParseV2$^*$} \cite{zhao2025boltzmann} & \rg{92.99} & \rg{94.36} & \rg{96.22} & \rg{84.63} & \rg{96.28} & \rg{78.18} & \rg{78.68} & \rg{88.76} & \rg{Not rep.} & \rg{Not rep.} & \rg{Not rep.} \\
\rg{TotalSegmentator$^*$} \cite{wasserthal2023totalseg} & \rg{92.10} & \rg{94.45} & \rg{94.17} & \rg{78.91} & \rg{95.61} & \rg{90.82} & \rg{78.10} & \rg{89.17} & \rg{Not rep.} & \rg{Not rep.} & \rg{Not rep.} \\
\midrule
\multicolumn{12}{l}{\textit{(a) Standalone zero shot VLM segmentation under matched training data (all models trained solely on 1,081 adult CT scans of TotalSegmentator)}} \\
\midrule
LISA$^\dagger$ \cite{lai2024lisa} & 86.43 & 88.12 & 89.75 & 67.89 & 93.02 & 71.78 & 54.29 & 78.76 & 73.01 & 27.03 & 583 / 809 \\
DenseMLLM$^\dagger$ \cite{li2026densemllm} & 87.07 & 89.48 & 92.09 & 69.66 & 92.97 & 79.36 & 58.77 & 81.34 & 75.43 & 30.20 & 494 / 873 \\
F-LMM$^\dagger$ \cite{wu2025flmm} & 90.15 & 91.27 & 93.54 & 73.18 & 94.47 & 83.34 & 68.84 & 84.97 & 81.80 & 23.00 & 472 / 606 \\
NativeSeg Round 0 (Ours) & 90.42 & 92.77 & 93.82 & 73.74 & 94.32 & 84.51 & 72.37 & 85.99 & 83.43 & 20.07 & 438 / 568 \\
\rowcolor{oursrow}\textbf{NativeSeg (Ours)} & \textbf{91.26} & \textbf{94.11} & \textbf{94.27} & \textbf{75.30} & \textbf{94.55} & \textbf{85.86} & \textbf{73.94} & \textbf{87.04} & \textbf{86.83} & \textbf{9.94} & \textbf{304} / \textbf{330} \\
\midrule
\multicolumn{12}{l}{\textit{(b) Plug and play native refinement across adult 2D nnU-Net and multi dataset 3D foundation models (overall Pediatric-CT-SEG SOTA)}} \\
\midrule
2D nnU-Net \cite{isensee2021nnunet} & 91.58 & 92.92 & 93.93 & 74.94 & 94.92 & 87.87 & 70.78 & 86.71 & 87.27 & 9.78 & 410 / 313 \\
2D nnU-Net + LRN$^\dagger$ \cite{chen2025lrn} & 87.90 & 87.10 & 88.82 & 63.78 & 94.54 & 78.00 & 60.72 & 80.12 & 65.99 & 158.51 & \textbf{37} / 12151 \\
2D nnU-Net + MedSAM2 \cite{ma2025medsam2} & 90.47 & \textbf{93.96} & \textbf{95.65} & \textbf{77.51} & 93.65 & 82.00 & 68.09 & 85.90 & 84.80 & 11.64 & 411 / \textbf{312} \\
\rowcolor{ourssub}\textbf{2D nnU-Net + NativeSeg (Ours)} & \textbf{91.59} & 93.20 & 93.97 & 75.55 & \textbf{94.94} & \textbf{87.90} & \textbf{70.86} & \textbf{86.86} & \textbf{87.49} & \textbf{9.71} & 398 / 313 \\
\cmidrule(lr){1-12}
3D SegVol$^\ddagger$ \cite{du2024segvol} & 88.95 & 89.00 & 92.46 & 65.22 & \textbf{95.75} & 86.13 & \textbf{76.89} & 84.91 & 86.01 & 10.05 & 720 / \textbf{167} \\
\rowcolor{ourssub}\textbf{3D SegVol$^\ddagger$ + NativeSeg (Ours)} & \textbf{90.15} & \textbf{91.22} & \textbf{93.67} & \textbf{72.57} & 95.72 & \textbf{87.43} & \textbf{76.89} & \textbf{86.81} & \textbf{88.05} & \textbf{8.81} & \textbf{551} / 175 \\
\cmidrule(lr){1-12}
3D VoxTell$^\ddagger$ \cite{rokuss2026voxtell} & \textbf{93.17} & 94.95 & 95.58 & \textbf{85.13} & \textbf{95.74} & 85.42 & 77.79 & 89.68 & 90.46 & 8.66 & 65 / 597 \\
3D VoxTell$^\ddagger$ + LRN$^\dagger$ \cite{chen2025lrn} & 88.48 & 87.93 & 90.47 & 69.17 & 95.23 & 79.03 & 63.74 & 82.01 & 67.63 & 158.01 & \textbf{20} / 12228 \\
3D VoxTell$^\ddagger$ + MedSAM2 \cite{ma2025medsam2} & 91.82 & \textbf{95.57} & \textbf{96.17} & 85.10 & 93.51 & 75.37 & 69.08 & 86.66 & 85.89 & 12.68 & 66 / 590 \\
\rowcolor{oursrow}\textbf{3D VoxTell$^\ddagger$ + NativeSeg (Ours)} & 93.12 & 95.13 & 95.61 & 84.76 & \textbf{95.74} & \textbf{86.69} & \textbf{77.88} & \textbf{89.85} & \textbf{90.53} & \textbf{8.28} & 67 / \textbf{562} \\
\bottomrule
\end{tabular}
\end{table*}

\paragraph{Datasets and Full Volume Evaluation Protocol.}
We train and evaluate in domain on publicly released, anonymized adult CT scans from \textbf{TotalSegmentator} \cite{wasserthal2023totalseg} ($1{,}081$ official training scans excluding corrupt scan \texttt{s0864}, $57$ validation, and $65$ test patients; \cref{tab:ts_main}) and evaluate zero shot out of distribution (OOD) transfer on anonymized pediatric CT scans from \textbf{Pediatric-CT-SEG} ($21$ patients; \cref{tab:pediatric}) \cite{pediatricctseg,jordan2022pediatric} across seven abdominal organs: spleen, right kidney, left kidney, gallbladder, liver, stomach, and pancreas.
Unlike slice benchmarks that prefilter positive slices using ground truth boxes, we evaluate every axial slice of every volume across all seven organ prompts, totaling $20{,}506$ axial slices ($143{,}542$ organ slice evaluations, $85.9\%$ target absent) on TotalSegmentator test and $5{,}366$ axial slices ($37{,}562$ organ slice evaluations: $7{,}102$ positive and $30{,}460$ absent, $81.1\%$ target absent) on Pediatric-CT-SEG.
Tables report per organ 3D DSC (averaged across target present patients) and their seven organ macro average alongside $3\,\text{mm}$ normalized surface Dice (NSD), conditional bidirectional HD95, missed positive slice evaluations (Miss$\downarrow$), and empty slice false positives (FP$\downarrow$); statistical tests report two sided paired Wilcoxon signed rank $p$ values over per patient macro DSC across target present patients ($n=56$ adult test, $n=21$ pediatric).

\paragraph{Leakage Audited and Controlled Backbone Comparisons.}
Evaluating multi dataset 3D foundation models on TotalSegmentator risks train on test overlap: documented pretraining lists of \textbf{3D SAT} \cite{zhao2025sat}, \textbf{BiomedParseV2} \cite{zhao2025biomedparse,zhao2025boltzmann}, \textbf{3D VoxTell} \cite{rokuss2026voxtell}, and \textbf{TotalSegmentator} \cite{wasserthal2023totalseg} overlap our adult test split ($53/65=81.5\%$ of test patients appear in SAT's training list, and VoxTell trained on TotalSegmentator).
We therefore adopt a two part benchmark design:
(i) On \textbf{TotalSegmentator} (\cref{tab:ts_main}), we compare \textbf{NativeSeg} against non native and single pass readout paradigms adapted to the \emph{identical} Qwen3.8-27B backbone and top $16$ layer LoRA configuration (\textbf{LISA}$^\dagger$ \cite{lai2024lisa}, \textbf{DenseMLLM}$^\dagger$ \cite{li2026densemllm}, and \textbf{F-LMM}$^\dagger$ \cite{wu2025flmm}), as well as fully supervised \textbf{2D nnU-Net} \cite{isensee2021nnunet} and external refiners (\textbf{LRN}$^\dagger$, a perturbation trained TotalSegmentator reimplementation following \cite{chen2025lrn}, and box prompted \textbf{MedSAM2} \cite{ma2025medsam2}).
(ii) On external \textbf{Pediatric-CT-SEG} (\cref{tab:pediatric}), we compare against both same backbone adult trained VLMs ($^\dagger$) and seven multi dataset foundation models (\textbf{BiomedParse}$^*$ \cite{zhao2025biomedparse}, \textbf{Text3DSAM}$^*$ \cite{xin2026text3dsam}, \textbf{3D SegVol}$^\ddagger$ \cite{du2024segvol}, \textbf{3D SAT}$^*$ \cite{zhao2025sat}, \textbf{BiomedParseV2}$^*$ \cite{zhao2025boltzmann}, \textbf{TotalSegmentator}$^*$ \cite{wasserthal2023totalseg}, and \textbf{3D VoxTell}$^\ddagger$ \cite{rokuss2026voxtell}) using fixed operating thresholds per mode ($\tau=0.05, K=3, \lambda_{\text{res}}=1.0$ in standalone mode; $\tau=0.10, K=2, \lambda_{\text{res}}=1.5$ in refiner mode; full training and modular evaluation details in \cref{sec:supp_repro}).

\section{Results and Analysis}
\label{sec:results}

\subsection{In Domain Performance on Adult CT}
\label{sec:res_adult}
\Cref{tab:ts_main} evaluates adult CT segmentation on the TotalSegmentator test split (full validation split in \cref{tab:ts_val}, \cref{sec:supp_val}).

\textbf{(a) Standalone Mode: Native Spatial Readout and Volumetric Refinement (\cref{tab:ts_main}(a)).}
Because prior VLM segmenters differ simultaneously in vision language backbones, pretraining corpora, and external mask heads, we evaluate representative readout designs on the identical Qwen3.8-27B backbone, LoRA rank, and adult training split.
Even in a single pass before any zooming, \textbf{NativeSeg Round $0$} ($86.01\%$ DSC) substantially outperforms non native 1D \texttt{[SEG]} broadcasting (\textbf{LISA}$^\dagger$, $80.25\%$), pure LLM spatial token decoding (\textbf{DenseMLLM}$^\dagger$, $82.22\%$), and single slice skip fusion (\textbf{F-LMM}$^\dagger$, $84.83\%$), confirming that keeping the 2D spatial token grid native to the VLM and coupling it with five slice CT context avoids 1D bottleneck compression.
Combining five slice Round $0$ readout with verified multiround local zoom and 3D axial continuity in \textbf{NativeSeg} lifts standalone accuracy to $86.81\%$ DSC ($p=2.3\times10^{-7}$ across $n=56$ target present patients, improving $48/56$ patients while cutting HD95 by $40.1\%$ from $13.13$ to $7.87\,\text{mm}$ and empty slice FPs by $31.3\%$), with the largest gains on small or deformable organs such as the gallbladder ($+3.18$ DSC) and stomach ($+1.02$ DSC).

\textbf{(b) Plug and Play Refiner Mode: Zero Degradation Specialist Editing (\cref{tab:ts_main}(b)).}
Paired with a strong in domain specialist (\textbf{2D nnU-Net}, $89.46\%$ DSC), a refiner must repair localized failures without corrupting already accurate masks.
Running external refiners \textbf{MedSAM2} \cite{ma2025medsam2} or perturbation trained \textbf{LRN} \cite{chen2025lrn} across every slice degrades \textbf{2D nnU-Net} to $87.37\%$ and $79.55\%$ DSC because repredicting masks without native error gating or cross model verification overwrites accurate boundaries.
By contrast, \textbf{2D nnU-Net + NativeSeg} halts at zero zooms on $96.3\%$ of evaluations (\cref{eq:stop}) and applies consensus verified updates (\cref{eq:verify}) selectively, preserving full specialist fidelity and nudging performance to $89.52\%$ DSC ($5.74\,\text{mm}$ HD95, $471$ FPs, with significant gains on challenging gallbladders at $p=0.018$).

\begin{figure*}[t]
\centering
\includegraphics[width=0.90\textwidth]{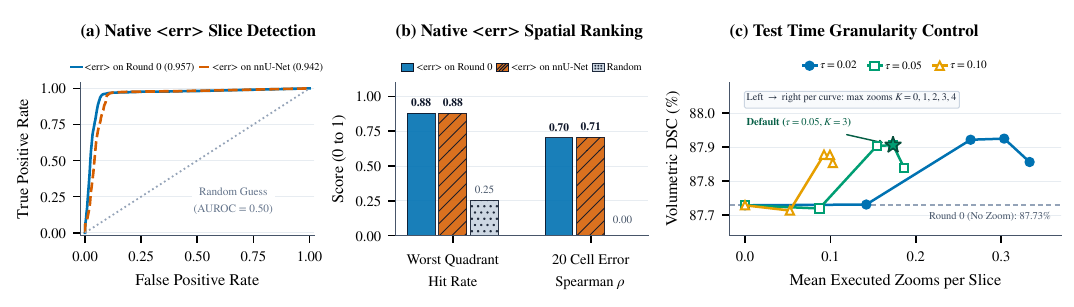}
\caption{Test time evaluation of \textbf{NativeSeg}'s native multiscale error gate ($\texttt{<err>}$) without ground truth on the validation ablation set. In both (a) and (b), the blue and vermilion series evaluate the \emph{same} frozen $\texttt{<err>}$ head when fed two different input mask sources: NativeSeg's own initial predictions (\emph{$\texttt{<err>}$ on Round $0$}, standalone mode) vs.\ unseen external specialist masks (\emph{$\texttt{<err>}$ on nnU-Net}, zero shot refiner mode; 2D nnU-Net itself has no error prediction head). (a) Receiver operating characteristic (ROC) curves and AUROC for flagging slices with $>5\%$ organ error ($200$ pixel floor). (b) Spatial error ranking capability of $\texttt{<err>}$ on slices with non zero error vs.\ \emph{Random} chance: \emph{Worst Quadrant Hit Rate} and \emph{20 Cell Error Spearman $\rho$} between predicted $E_{1:20}^{(0)}$ and true $e_{1:20}^{(0)}$. (c) Standalone accuracy vs.\ mean executed zooms per slice across tolerances $\tau\in\{0.02,0.05,0.10\}$ ($K=0,1,2,3,4$ from left to right).}
\label{fig:error_pareto}
\end{figure*}

\subsection{Zero Shot Generalization on Pediatric CT}
\label{sec:res_ped}
\Cref{tab:pediatric} evaluates zero shot transfer to external Pediatric-CT-SEG alongside multi dataset foundation models trained on large scale extra corpora (gray reference block).

\textbf{(a) Matched Data Standalone SOTA and Cross Tier Comparison (\cref{tab:pediatric}(a)).}
Pediatric CT introduces pronounced shifts in body size, organ scale, and visceral fat contrast relative to adult training scans while providing an external benchmark disjoint from TotalSegmentator pretraining.
Under matched training data ($1{,}081$ adult CT scans only), standalone \textbf{NativeSeg} achieves state of the art accuracy across every organ and metric ($87.04\%$ DSC, $p=6.7\times10^{-6}$ over Round $0$, cutting HD95 from $20.07$ to $9.94\,\text{mm}$), surpassing non native 1D \texttt{[SEG]} readout (\textbf{LISA}$^\dagger$, $78.76\%$) by $+8.28$ DSC, single pass 2D VLMs, and adult \textbf{2D nnU-Net} ($86.71\%$).
Moreover, despite training on an order of magnitude fewer scans than the external multi dataset reference block, standalone \textbf{NativeSeg} already outperforms \textbf{BiomedParse}$^*$ ($21.19\%$), \textbf{Text3DSAM}$^*$ ($42.64\%$), and \textbf{3D SegVol}$^\ddagger$ ($84.91\%$, reducing missed positive slices by $57.8\%$ from $720$ to $304$).
\begin{table}[!t]
\centering
\scriptsize
\renewcommand{\arraystretch}{0.76}
\setlength{\tabcolsep}{1.6pt}
\caption{Ablation studies on the fixed 6 patient TotalSegmentator validation subset (14,049 organ slice evaluations; $^\dagger$controlled readout on Qwen3.8-27B; \emph{indiv.}\ denotes individual additions to \emph{Error Gate + Verify} on the frozen single slice refiner; DSC/NSD in \%, HD95 in mm; \cref{sec:supp_repro}).}
\label{tab:ablations}
\begin{tabular}{lrrrr}
\toprule
Variant (Qwen3.8-27B) & DSC $\uparrow$ & NSD $\uparrow$ & HD95 $\downarrow$ & Miss$\downarrow$/FP$\downarrow$ \\
\midrule
\multicolumn{5}{l}{\textit{(a) Round 0 Mask Readout: Non Native vs.\ Native Spatial}} \\
Discrete Codebook (MedSIGHT$^\dagger$ \cite{chang2026medsight}) & 68.91 & 60.84 & 48.9 & \textbf{56}/1044 \\
Single \texttt{[SEG]} Broadcast (LISA$^\dagger$ \cite{lai2024lisa}) & 81.68 & 76.27 & 21.4 & 136/140 \\
Spatial Tokens Only (DenseMLLM$^\dagger$ \cite{li2026densemllm}) & 83.73 & 79.54 & 19.1 & 153/119 \\
Spatial Tokens + Skip (F-LMM$^\dagger$ \cite{wu2025flmm}) & 87.73 & 84.64 & \textbf{16.3} & 134/116 \\
\rowcolor{oursrow}\textbf{Spatial Tokens + Five Slice U-Net} & \textbf{88.72} & \textbf{86.39} & 17.0 & 113/\textbf{97} \\
\midrule
\multicolumn{5}{l}{\textit{(b) Native Multiround Control \& Component Analysis}} \\
$1\times$ Global Reread (No Zoom) & 86.17 & 82.42 & 21.2 & 69/239 \\
Exhaustive $2\times2$ Sweep \cite{cheng2020cascadepsp} & 86.21 & 78.47 & 57.9 & \textbf{68}/257 \\
No Mask Conditioning ($\alpha_{\text{ov}}=0$) & 86.82 & 80.19 & 25.4 & 134/116 \\
Pixel Entropy Selection \cite{zenk2025failure} & 87.11 & 82.99 & 24.6 & 78/212 \\
Random Region Selection & 87.25 & 81.63 & 48.0 & 89/196 \\
\cmidrule(lr){1-5}
Error Gate (No Verify, \cref{eq:verify}) & 87.28 & 84.03 & 16.6 & 134/115 \\
+ Verify \& Freeform ROI (indiv.) & 87.85 & 84.78 & 16.2 & 134/115 \\
+ Confidence Map Memory (indiv.) & 87.91 & 84.86 & 16.3 & 134/114 \\
+ 3D Axial Continuity (indiv.) & 88.54 & 85.75 & 15.5 & 97/145 \\
\rowcolor{oursrow}\textbf{NativeSeg (+ Joint Stage B)} & \textbf{89.09} & \textbf{88.40} & \textbf{7.1} & 115/\textbf{52} \\
\bottomrule
\end{tabular}
\end{table}

\begin{figure}[t]
\centering
\includegraphics[width=0.91\columnwidth]{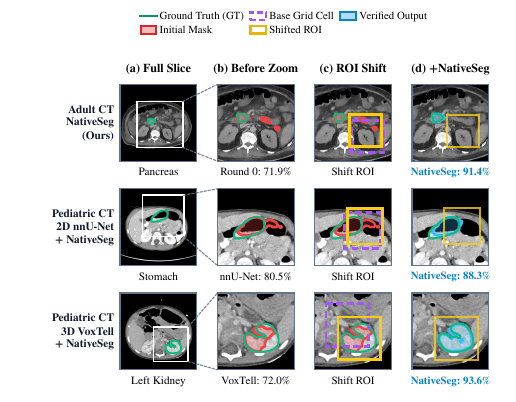}
\caption{Qualitative examples in standalone mode (top: adult pancreas) and specialist refiner mode (middle: pediatric stomach with \textbf{2D nnU-Net}; bottom: pediatric left kidney with \textbf{3D VoxTell}). Column (a) shows the full CT slice with a white magnifier window enlarged for visualization across (b) to (d): (b) initial full organ mask before zoom, (c) fixed $4\times4$ base grid cell vs.\ shifted local ROI targeting the boundary error, and (d) final \textbf{NativeSeg} mask after pasting the local update into (b) (\cref{eq:blend}: pixels outside the yellow shifted ROI remain identical to (b), while all boundary edits occur strictly inside the shifted ROI; full slice DSC below each panel).}
\label{fig:qualitative}
\end{figure}

\textbf{(b) Complementary Specialist Refinement for Overall Benchmark SOTA (\cref{tab:pediatric}(b)).}
While massive multi dataset 3D models (\textbf{3D SAT}$^*$ $88.01\%$, \textbf{BiomedParseV2}$^*$ $88.76\%$, \textbf{TotalSegmentator}$^*$ $89.17\%$, and \textbf{3D VoxTell}$^\ddagger$ $89.68\%$) leverage tens of thousands of extra training volumes, \textbf{NativeSeg} complements rather than merely competes with them: whereas external refiners \textbf{LRN}$^\dagger$ and \textbf{MedSAM2} degrade all base models (losing $3.02$ to $7.67$ DSC on \textbf{3D VoxTell}$^\ddagger$), frozen adult trained \textbf{NativeSeg} halts at zero zooms on $93.8\%$ to $96.5\%$ of evaluations and lifts \textbf{2D nnU-Net} to $86.86\%$ ($p=0.024$), \textbf{3D SegVol}$^\ddagger$ to $86.81\%$ ($+1.90$ DSC, $p=9.5\times10^{-6}$), and \textbf{3D VoxTell}$^\ddagger$ to $89.85\%$ DSC ($90.53\%$ NSD, $8.28\,\text{mm}$ HD95)---surpassing every standalone 2D and 3D foundation model in \cref{tab:pediatric} (\cref{fig:qualitative}).

\subsection{Ablation Studies and Granularity Control}
\label{sec:res_ablate}

\textbf{(a) Why Native Spatial Readout Matters (\cref{tab:ablations}(a)).}
Comparing five Round $0$ mask readouts on Qwen3.8-27B (\cref{sec:supp_repro}) reveals why non native bottlenecks hurt CT delineation: discrete codebook quantization (\textbf{MedSIGHT}$^\dagger$ \cite{chang2026medsight}, $68.91\%$) suffers severe blockiness and 1D \texttt{[SEG]} broadcasting (\textbf{LISA}$^\dagger$ \cite{lai2024lisa}, $81.68\%$) loses fine curvature, whereas reading masks natively from 2D spatial tokens (\textbf{DenseMLLM}$^\dagger$, $83.73\%$), fusing $32\times32$ vision skips (\textbf{F-LMM}$^\dagger$, $87.73\%$), and incorporating five slice U-Net decoding ($88.72\%$) progressively resolve spatial and axial ambiguity.

\textbf{(b) Native Closed Loop Control vs.\ Heuristic Policies (\cref{tab:ablations}(b)).}
Starting from identical single slice Round $0$ masks ($87.73\%$) on one frozen zoom checkpoint, $1\times$ global rereading ($86.17\%$) lowers DSC, confirming gains require higher local spatial resolution rather than extra forward depth; exhaustive $2\times2$ sweeping \cite{cheng2020cascadepsp}, random cropping, and pixel entropy routing \cite{zenk2025failure} degrade masks via unverified overediting, while zeroing mask conditioning drops DSC to $86.82\%$.
Adding verification with freeform ROI shifting ($87.85\%$), continuous confidence map memory ($87.91\%$), or axial continuity ($88.54\%$) individually improves the frozen single slice refiner, and uniting them in the joint five slice Stage B checkpoint reaches $89.09\%$ DSC and $7.1\,\text{mm}$ HD95 on the validation subset.

\textbf{(c) Native Error Gate Calibration and Compute Control (\cref{fig:error_pareto}).}
Without ground truth or external quality control networks, $\texttt{<err>}$ natively governs \emph{which slices} to refine (\cref{fig:error_pareto}(a)), \emph{where} to zoom (\cref{fig:error_pareto}(b)), and \emph{when} to stop (\cref{fig:error_pareto}(c)).
Crucially, both the blue and vermilion series in \cref{fig:error_pareto}(a, b) evaluate our \emph{same} frozen native $\texttt{<err>}$ head across two input mask sources---NativeSeg's own Round $0$ masks (standalone mode) and unseen \textbf{2D nnU-Net} masks (zero shot refiner mode)---benchmarking $\texttt{<err>}$'s error detection and spatial ranking ability against random chance rather than comparing against an external detector.
Across all $14{,}049$ validation evaluations, $\texttt{<err>}$ achieves $0.957$ AUROC on Round $0$ masks and transfers zero shot at $0.942$ AUROC to unseen \textbf{2D nnU-Net} masks (\cref{fig:error_pareto}(a)).
On slices with non zero error (\cref{fig:error_pareto}(b)), $\texttt{<err>}$ ranks spatial boundary errors with zero transfer degradation between its own Round $0$ masks and unseen \textbf{2D nnU-Net} masks, reaching an $0.88$ worst quadrant hit rate on both ($3.5\times$ the $0.25$ random baseline) and $0.70$ / $0.71$ per slice Spearman $\rho$ across all $20$ multiscale cells, directly explaining why NativeSeg refines external specialists zero shot without retraining.
Sweeping $(\tau, K)$ at test time (\cref{fig:error_pareto}(c)) prevents $K=4$ overrefinement and locates the Pareto knee at $\tau=0.05, K=3$ for standalone mode ($0.17$ mean zooms/slice on the validation sweep; $0.14$ to $0.16$ on full splits) and $\tau=0.10, K=2$ for refiner mode.

\section{Conclusion}
\label{sec:discussion}

We introduced \textbf{NativeSeg}, demonstrating that dense 3D CT mask readout, empty slice screening, spatial error diagnosis, and closed loop local refinement can be performed \emph{natively} over a vision language model's 2D spatial visual tokens without external SAM decoders, discrete mask codebooks, or stateless box prompting tools.
By uniting five slice spatial token decoding and presence gating ($\texttt{<pres>}$) with multiscale error gated local zoom ($\texttt{<err>}$), closed loop verification, and 3D axial continuity on LoRA adapted Qwen3.8-27B, NativeSeg substantially outperforms non native VLM readouts in domain and zero shot on pediatric CT while safely refining 2D and 3D specialists.

{\small
\bibliographystyle{ieeenat_fullname}
\bibliography{refs}
}

\clearpage
\maketitlesupplementary
\appendix
\raggedbottom
\setlength{\dbltextfloatsep}{10pt plus 2pt minus 2pt}
\section{Scope and Operational Considerations}
\label{sec:supp_limitations}

\paragraph{Evaluated Anatomical Scope.}
Our evaluation trains on $1{,}081$ official adult CT volumes (excluding corrupt scan \texttt{s0864}) and evaluates on $65$ test and $57$ validation volumes from TotalSegmentator \cite{wasserthal2023totalseg} plus $21$ external pediatric CT volumes from Pediatric-CT-SEG \cite{pediatricctseg,jordan2022pediatric} ($1{,}224$ volumes total) across seven abdominal organs (spleen, right/left kidney, gallbladder, liver, stomach, and pancreas), matching recent volumetric benchmarks \cite{du2024segvol,zhao2025sat,rokuss2026voxtell}.
Although all native modules of NativeSeg (2D spatial token readout, 20 cell error gating, local zoom, and verification) are shared across organs without organ specific heads, extending evaluation to thoracic, vascular, musculoskeletal, and lesion targets is an important future direction.

\paragraph{Local 2.5D Axial Window Versus Full Volume Memory.}
To balance in plane resolution ($512\times512$ canvas) with volumetric efficiency, NativeSeg encodes a five slice stack $[I_{z-2},\dots,I_{z+2}]$ per slice, screens empty slices via $\texttt{<pres>}$, and applies deterministic axial continuity and 3D connected component filtering across stacked slice predictions, leaving end to end full volume 3D token memory open for future work.

\paragraph{Execution Budget and Backbone Memory Footprint.}
Across full volume benchmarks, NativeSeg halts at zero executed zoom rounds on $91.8\%$ to $93.0\%$ of standalone organ slice evaluations ($0.14$ to $0.16$ mean zoom rounds/slice; $0.17$ on the $6$ patient validation sweep) and on $93.8\%$ to $96.5\%$ of specialist refiner evaluations; a single Round $0$ pass averages $0.26\,\text{s}$ per evaluation (batch size $1$) on one NVIDIA H20 GPU.
Storing the shared frozen Qwen3.8-27B backbone in bfloat16 requires $\sim 54\,\text{GB}$ of weight memory ($\sim 58\,\text{GB}$ peak GPU memory per loaded LoRA/decoder state: $56.0\text{M}$ trainable parameters for a five slice checkpoint and $34.6\text{M}$ for a single slice checkpoint); low bit quantization of the frozen lower $48$ layers is a natural next step.

\section{Reproducibility and Ablation Details}
\label{sec:supp_repro}

\paragraph{Training Configuration, Modular Pipeline, and Volumetric Continuity.}
All Qwen3.8-27B models in \Cref{tab:ts_main,tab:pediatric,tab:ablations,tab:ts_val} share the same frozen vision tower, frozen lower $48$ language layers, and rank $16$ LoRA adapters ($\alpha_{\text{LoRA}}=32$, $<0.3\%$ of backbone parameters; $29.2\text{M}$ LoRA parameters across Layers $49$ to $64$, $1.4\text{M}$ spatial offsets and gate heads, plus $25.4\text{M}$ parameters in the five slice U-Net decoder or $4.0\text{M}$ in the single slice skip decoder).
Models train with deterministic seeds, organ balanced slice sampling ($50\%$ target present, $50\%$ absent in Stage A; $70\%$ positive in Stage B, with hard organ oversampling on pancreas, gallbladder, stomach, and kidneys for five slice runs), random scaling in $[0.75,1.25]$ (probability $0.5$), and AdamW (weight decay $10^{-2}$, $200$ warmup steps, cosine decay, LoRA learning rate $10^{-4}$ to $2\times10^{-4}$, decoder/head learning rate $2\times10^{-4}$ to $4\times10^{-4}$).
Single slice readout baselines train for $5{,}000$ steps (\texttt{[SEG]} broadcast and spatial tokens only) to $12{,}000$ steps (spatial tokens + skip; $20{,}000$ steps for discrete codebook), followed by $3{,}000$ steps for the single slice multiround local zoom refiner; initialized from this spatial representation, the five slice Stage A readout trains for $6{,}000$ steps under \cref{eq:loss} and the joint five slice Stage B checkpoint trains for $4{,}000$ steps across clean ($15\%$ prior), perturbed ($35\%$), Round $0$, and extra zoom states.
For full split standalone benchmarks (\Cref{tab:ts_main,tab:pediatric,tab:ts_val}), single pass Round $0$ baselines report presence gated slice predictions ($p_z\ge 0.5$), whereas modular multiround inference pairs five slice Stage A Round $0$ proposals $(B_z, p_z^{\text{R0}})$ with verified zoom updates $(M_z^{\text{zm}}, p_z^{\text{zm}})$ from the local refiner initialized on the single slice spatial branch ($\tau=0.05, K=3, \lambda_{\text{res}}=1.0$): blending $p_z = 0.65\,p_z^{\text{R0}} + 0.35\,p_z^{\text{zm}}$ and $(B_z \wedge \mathbb{1}[p_z^{\text{R0}}\ge 0.5]) \vee (M_z^{\text{zm}} \wedge B_z)$, followed by two pass axial tip recovery ($p_z \ge 0.25$ overlapping active neighbors), single slice gap bridging, isolated slice pruning ($p_z < 0.62$ without neighbor support $\ge 0.35$), and 3D component filtering (retaining components with voxel count $\times$ mean presence $\ge 0.22$ of the maximum component).
Slice masks are scored in original $(H,W)$ space with axial spacing to compute per organ 3D DSC (macro averaged across seven organs), $3\,\text{mm}$ NSD, conditional HD95, and paired two sided Wilcoxon $p$ values over per patient macro DSC across target present patients ($n=56$ adult test; $n=21$ pediatric).

\begin{table*}[t]
\centering
\scriptsize
\renewcommand{\arraystretch}{0.78}
\setlength{\tabcolsep}{3.5pt}
\caption{Full validation split segmentation performance on TotalSegmentator (57 patients, 14,095 axial slices, 98,665 organ slice evaluations; $^\dagger$controlled readout paradigm adapted to Qwen3.8-27B and trained on the 1,081 patient training split; Spl: spleen, RKid/LKid: right/left kidney, Gall: gallbladder, Liv: liver, Sto: stomach, Pan: pancreas).}
\label{tab:ts_val}
\begin{tabular}{lrrrrrrrrrrr}
\toprule
Method & Spl $\uparrow$ & RKid $\uparrow$ & LKid $\uparrow$ & Gall $\uparrow$ & Liv $\uparrow$ & Sto $\uparrow$ & Pan $\uparrow$ & Mean DSC (\%) $\uparrow$ & NSD (\%) $\uparrow$ & HD95 (mm) $\downarrow$ & Miss $\downarrow$ / FP $\downarrow$ \\
\midrule
\multicolumn{12}{l}{\textit{(a) Standalone VLM segmentation on the identical Qwen3.8-27B backbone (Validation Split, 57 patients)}} \\
\midrule
LISA$^\dagger$ \cite{lai2024lisa} & 89.46 & 87.88 & 83.79 & 63.75 & 92.65 & 76.52 & 56.60 & 78.66 & 74.38 & 29.05 & 1057 / 898 \\
DenseMLLM$^\dagger$ \cite{li2026densemllm} & 89.69 & 87.64 & 86.14 & 64.31 & 93.40 & 80.87 & 62.08 & 80.59 & 77.61 & 25.53 & 1006 / 881 \\
F-LMM$^\dagger$ \cite{wu2025flmm} & 92.90 & 90.68 & 87.95 & 67.44 & 95.19 & 83.96 & 69.74 & 83.98 & 82.83 & 19.45 & 940 / 713 \\
NativeSeg Round 0 (Ours) & 93.31 & 92.47 & 90.63 & 69.96 & 95.24 & 85.24 & 72.73 & 85.65 & 84.79 & 18.95 & 712 / 728 \\
\rowcolor{oursrow}\textbf{NativeSeg (Ours)} & \textbf{94.37} & \textbf{93.25} & \textbf{91.45} & \textbf{73.44} & \textbf{95.45} & \textbf{86.21} & \textbf{73.46} & \textbf{86.80} & \textbf{87.67} & \textbf{10.15} & \textbf{576} / \textbf{475} \\
\midrule
\multicolumn{12}{l}{\textit{(b) Plug and play refinement of 2D nnU-Net (Validation Split, 57 patients)}} \\
\midrule
2D nnU-Net \cite{isensee2021nnunet} & 95.38 & \textbf{95.13} & 92.46 & 74.09 & 96.76 & 89.78 & 80.46 & 89.15 & 91.81 & \textbf{8.65} & 456 / 496 \\
2D nnU-Net + LRN$^\dagger$ \cite{chen2025lrn} & 91.10 & 87.39 & 76.27 & 56.87 & 95.43 & 82.15 & 65.04 & 79.18 & 66.79 & 188.15 & \textbf{61} / 32294 \\
2D nnU-Net + MedSAM2 \cite{ma2025medsam2} & 93.99 & 93.17 & 91.32 & 73.39 & 94.07 & 86.95 & 76.01 & 86.99 & 87.92 & 11.59 & 456 / 496 \\
\rowcolor{ourssub}\textbf{2D nnU-Net + NativeSeg (Ours)} & \textbf{95.40} & \textbf{95.13} & \textbf{92.69} & \textbf{74.27} & \textbf{96.77} & \textbf{89.83} & \textbf{80.55} & \textbf{89.23} & \textbf{91.86} & 8.69 & 456 / \textbf{488} \\
\bottomrule
\end{tabular}
\end{table*}

\paragraph{Plug and Play Specialist Refinement Protocol.}
Complementing \cref{sec:method,sec:res_adult}, when NativeSeg refines binary masks $M_0^{\text{spec}}$ from an external model (2D nnU-Net, 3D SegVol, or 3D VoxTell), the local zoom refiner ($\tau=0.10, K=2, \lambda_{\text{res}}=1.5$) inspects $O_0^{(0)}$ and halts at zero zooms whenever $E_{c^\star}^{(0)} / \max(s_{\text{canvas}}^2 |M_0^{\text{spec}}|, a_{\min}) < \tau$ ($93.8\%$ to $96.5\%$ of evaluations).
Where verified edits are proposed (\cref{eq:verify}), our parameter free cross model consensus filter retains pixelwise additions agreed upon by both the verified zoom output and standalone five slice Round $0$ proposal ($p_z^{\text{R0}} \ge 0.5$), retains agreed background removals, and prunes isolated single slice specialist false positives when both $z-1$ and $z+1$ are empty in $M_0^{\text{spec}}$ and $p_z^{\text{R0}} < 0.03$ with an empty Round $0$ proposal.
External baselines \textbf{LRN}$^\dagger$ (perturbation trained CNN refiner on TotalSegmentator \cite{chen2025lrn}) and \textbf{MedSAM2} \cite{ma2025medsam2} (box prompted per active slice of $M_0^{\text{spec}}$) are evaluated on the exact same organ slice evaluations.

\paragraph{Ablation Variant Procedures (\Cref{tab:ablations}).}
All variants in \Cref{tab:ablations} and \cref{fig:error_pareto} use the fixed $6$ patient TotalSegmentator validation subset ($2{,}007$ slices, $14{,}049$ evaluations: $2{,}247$ positive, $11{,}802$ absent):

\noindent\textit{Part (a): Round $0$ Mask Readout Architecture (\cref{tab:ablations}(a)).}
All five variants share the Qwen3.8-27B backbone and top $16$ layer LoRA configuration, varying the Round $0$ mask readout design:
\begin{itemize}
\item \textbf{Discrete Codebook (MedSIGHT$^\dagger$ \cite{chang2026medsight})}: Predicts $32\times32$ indices from a learned $512$ entry vector quantized codebook decoded to original slice resolution.
\item \textbf{Single \texttt{[SEG]} Broadcast (LISA$^\dagger$ \cite{lai2024lisa})}: Broadcasts one 1D Layer $64$ token across $32\times32$ and fuses it with $F_{V_0}$, isolating the non native 1D \texttt{[SEG]} bottleneck.
\item \textbf{Spatial Tokens Only (DenseMLLM$^\dagger$ \cite{li2026densemllm})}: Decodes the $256$ Layer $64$ 2D spatial token states $H_{V_0}$ ($16\times16$ grid) natively without vision skip features.
\item \textbf{Spatial Tokens + Skip (F-LMM$^\dagger$ \cite{wu2025flmm})}: Fuses $16\times16$ states $H_{V_0}$ with $32\times32$ patch features $F_{V_0}$ in a single slice decoder without five slice context.
\item \textbf{Spatial Tokens + Five Slice U-Net (Ours)}: Uses our five stage residual U-Net decoder conditioned on $[I_{z-2},\dots,I_{z+2}]$ and $[-60,140]\,\text{HU}$.
\end{itemize}

\noindent\textit{Part (b): Multiround Control and Individual Component Ablations (\cref{tab:ablations}(b)).}
To control for training variance, the first nine rows of \cref{tab:ablations}(b) start from the \emph{exact same} single slice Round $0$ predictions (\texttt{Spatial Tokens + Skip}, $87.73\%$ DSC) on one frozen $3{,}000$ step single slice zoom checkpoint, testing heuristic policies and individual verified components before full joint five slice Stage B training:
\begin{itemize}
\item \textbf{$1\times$ Global Reread (No Local Zoom)}: Reprocesses full $512\times512$ $O_0^{(t)}$ at $1\times$ scale for $K=2$ rounds without cropping, isolating local spatial resolution from pass depth.
\item \textbf{Exhaustive $2\times2$ Patch Sweep \cite{cheng2020cascadepsp}}: Refines all four $2\times2$ quadrants on positive slices without $\texttt{<err>}$ halting or verification.
\item \textbf{Without Mask Conditioning ($\alpha_{\text{ov}}=0$, Blank Prior)}: Zeros out both the visual RGB mask overlay and decoder prior state channel on $O_0^{(t)}$ and $O_{r^\star}^{(t)}$.
\item \textbf{Pixel Entropy Selection \cite{zenk2025failure}}: Selects the unvisited cell of highest pixelwise Shannon entropy of $\sigma(L_0)$ instead of $\texttt{<err>}$.
\item \textbf{Random Region Selection}: Uniformly samples unvisited cells $c\in\{1,\dots,20\}$ for $K=2$ rounds.
\item \textbf{Error Gate (Without Verification, \cref{eq:verify})}: Uses $\texttt{<err>}$ for halting (\cref{eq:stop}) and cell selection while accepting every zoom update unconditionally.
\item \textbf{+ Verification \& Freeform ROI (indiv., \cref{eq:shift,eq:verify})}: Individually enables error verification (\cref{eq:verify}), freeform ROI shifting (\cref{eq:shift}), and border feathering (\cref{eq:blend}) on the frozen checkpoint.
\item \textbf{+ Confidence Map Memory (indiv., \cref{eq:residual,eq:blend})}: Individually enables verification with continuous logit retention across zoom rounds on the frozen checkpoint.
\item \textbf{+ 3D Axial Continuity (indiv.)}: Individually enables verification with axial continuity and 3D component filtering on the frozen checkpoint.
\item \textbf{NativeSeg (+ Joint Stage B, Ours)}: Unites five slice U-Net readout, freeform ROI feathering, continuous logit memory, and 3D axial continuity in one $4{,}000$ step five slice Stage B checkpoint ($89.09\%$ Val6 DSC; $85.83\%$ adult test and $86.54\%$ pediatric DSC as a single checkpoint vs.\ $86.81\%$ and $87.04\%$ stage decoupled).
\end{itemize}

\section{Additional Quantitative Results on the Validation Split}
\label{sec:supp_val}

\Cref{tab:ts_val} reports complete per organ and volumetric results on the full TotalSegmentator validation split ($57$ patients, $14{,}095$ axial slices, $98{,}665$ organ slice evaluations), comparing \textbf{NativeSeg} against matched backbone VLM readout paradigms (\texttt{[SEG]} broadcast \cite{lai2024lisa} and 2D spatial token decoding \cite{wu2025flmm,li2026densemllm}) as well as external specialist refiners (\textbf{LRN}$^\dagger$ \cite{chen2025lrn} and \textbf{MedSAM2} \cite{ma2025medsam2}).

\end{document}